# Metric Learning with Dynamically Generated Pairwise Constraints for Ear Recognition

Ibrahim Omara[a,b], Hongzhi Zhang[a], Faqiang Wang[a], Wangmeng Zuo[a]

[a] School of Computer Science and Technology, Harbin Institute of Technology, Harbin 150 0 01, P.R. China

[b] Department of Mathematics, Faculty of Science, Menoufia University, Shebin El-kom, 32511, Egypt

i_omara84@yahoo.com, zhanghz0451@gmail.com, tshfqw@163.com, cswmzuo@gmail.com

**Abstract**. Ear recognition task is known as predicting whether two ear images belong to the same person or not. In this paper, we present a novel metric learning method for ear recognition. This method is formulated as a pairwise constrained optimization problem. In each training cycle, this method selects the nearest similar and dissimilar neighbors of each sample to construct the pairwise constraints, and then solve the optimization problem by the iterated Bregman projections. Experiments are conducted on AMI, USTB II and WPUT databases. The results show that the proposed approach can achieve promising recognition rates in ear recognition, and its training process is much more efficient than the other competing metric learning methods.

## 1 Introduction

As an important person authentication technique, biometric recognition has been widely applied in surveillance applications, forensics and criminal investigations. Since the biometric traits are unique, universal and permanent, biometric recognition is more secure and reliable than the traditional person authentication approaches.

Among the existing biometric recognition methods, there have been many kinds of biometric traits, e.g. face, finger-print, palm-print, iris, signature, voice, key-stroke and gait. Compared with the other traits, human ear has a stable structure with different ages [1]. Also, the ear is insensitive to the variations such as make-up, glasses, and facial expression [2]. The ear image is also easy to acquire with little person awareness and user cooperation [3]. Furthermore, it has been proven that the left and right ears of the same person have some similarities, but are not strictly symmetric [4]. Therefore, ear recognition has received increasing research interest.

For the ear recognition task, one of the most common methods is the metric learning method, which aims to learn the distance between two instances, where the distances between similar instances are shorter than those between dissimilar instances. It plays a crucial role in machine learning and successfully applied into many biometric recognition tasks. Most of the existing metric learning methods learn the distance metric from the pairwise or triplet constraints [5]. The pairwise constraints make the distances of similar pairs shorter than a given threshold, while the distances of dissimilar pairs longer than this threshold. The triplet constraints make the distance of similar samples shorter than that of dissimilar samples. Many existing metric learning methods learn the distance metric from the pairwise constraints [5], which make the distances of similar pairs shorter than a given threshold, and the distances of dissimilar pairs longer than this threshold. As the quantity of pairwise constraints is very large ($O(N^2)$ pairs can be constructed from $N$ samples), the existing metric learning methods usually select part of the pairwise constraints for training. Davis et al. [6] proposed to select the pairwise constraints randomly. Wang et al. [7] propose a strategy to construct the pairs from the training samples. For each training sample, its nearest similar and dissimilar samples are used to construct the similar and dissimilar pairs.

The previous metric learning methods construct the pairwise constraints as a preprocessing step, and use the fixed pairwise constraints in training. This strategy, however, suffers from evident drawbacks. As the number of training pairs is limited, and some pairs are never used in training, the trained model will under-fits the non-used training pairs. To address the aforementioned limitation, we propose a novel method to learn the distance metric from online generated pairwise constraints for ear recognition. First, it extracts the local phase quantization (LPQ), histogram of oriented gradient (HOG), and Gabor features of ear images. Then it uses the Discriminant Correlation Analysis (DCA) method to fuse different features and reduce the feature dimension. Finally it learns the distance metric based on the extracted feature. In this method, we learn the distance metric for several cycles. In each cycle, we construct the pairwise constraints by the trained distance in last cycle, and learn the distance metric based on these pairwise constraints. As the pairwise constraints are updated in each cycle, the training pairs in our proposed method are more than those in the previous metric learning methods. We conduct the experiments on several ear image datasets to evaluate our proposed method. The results show that our proposed method outperforms the state-of-the-art metric learning methods in ear recognition.

The rest of this paper is organized as follows: Section 2 introduces the related work about the commonly used ear recognition methods and metric learning algorithms. Section 3 demonstrates the proposed approach based on metric learning. Section 4 presents the experimental results and discussion. Finally, Section 5 draws the conclusion of this paper.

## 2 Related Work

In this section, we give a brief review on the related works from two aspects, i.e. ear recognition and metric learning.

### 2.1 Ear recognition

The existing works on ear recognition mainly focus on two aspects, i.e. feature extraction and classification. Two kinds of features, i.e. geometric and appearance-based features, are mainly used in the ear recognition methods. The geometric features include maximum ear height line (EHL) [8, 9], inner and outer helixes [8], tragus [10], etc. Some method uses the combination of these features [11, 12] which can facilitate the discriminability of the geometric features. For the appearance-based features, it mainly includes intensity, directional and spatial-temporal information. Many of works have presented holistical features such as Eigenear and Eigenface [2], ICA [13], active shape model to detect outer ear contour [14], 1 D and 2 D Gabor filter [15, 16], and locale features such as LBP [17], HOG [18], kernel of polar sine transform (PST) [19], SIFT [20], and SURF [21]. Recently, Nanni and Lumini [22] adopted the sequential forward floating selection (SFFS) to select the best features from sub-windows in an ear image. Yuan and Mu [23] presented a brief review of ear recognition and proposed a fusion method for ear recognition based on local information. Most of known ear recognition methods adopt the approximate nearest neighbor (ANN) [19] or support vector machine (SVM) [24, 25] as the classifier. In recent years, the RBF [13], neural networks [26], and pairwise SVM [27] have also been applied into ear recognition. For more details, please refer to [28, 29] for the comprehensive surveys on ear recognition methods.

We can notice that, famous and typical ear recognition methods for 2D ear images focus on the feature extraction process, and adopt a simple nearest neighbor, or typical SVM methods for ear recognition problem. Different distances are mainly adopted as the matching criterion, such as Euclidean distance [11, 30, 31], Hamming distance [32, 10, 9], RBF[13], BP [8] and neural networks [26] and recently, discriminative classifiers, such as ANN [19], SVM [24, 25], and pairwise SVM have drawn much attention [27]. However, due to the lack of training images of ears and the multiple class property of ear database, typical matching processes may not lead to the desired performance. Whereas, Mahalanobis matrix takes the correlation of various features as the elements of the off-diagonal, and it is scale invariant. Therefore, appropriate similarity distance metric should be considered to our proposed method, called a Mahalanobis distance metric, it learning over prior information to measure the similarity or dissimilarity between different instances. Therefore, we investigate an alternative method for ear recognition problem; metric learning extends the similarity measurement to take the advantage of prior information as a label over standard similarity measures. Metric learning has been proposed to learn Mahalanobis distance metrics for k-nearest neighbor classification and has better interclass generalization performance, which can be applied to handle the multiple classes of ear image database better.

## 2.2 Metric Learning

Metric learning plays a crucial role for various applications such as signature verification [33], data classification [34], and person re-identification [35]. The existing metric learning can be divided into two categories, i.e. pairwise constrained metric learning and triplet constrained metric learning. The pairwise constrained metric learning methods include ITML [6], LDML [36], DML-eig [37], and SML [38], etc. The ITML method formulates the problem as minimizing the LogDet divergence instance between the learned distance metric and prior distance metric subject to the pairwise constraints [6]. It can be solved by the iterated Bregman projection algorithm. Guilllaumin et al. [36] proposed the LDML method. It defines the probabilities of each sample pair to be similar and dissimilar, respectively, and formulates the problem as maximizing the log-likelihood of all training pairs. Ying et al. proposes the DML-eig method by formulating metric learning as an eigenvalue optimization problem with pairwise constraints [37]. Kostinger et al. proposes the KISSME method to learn the distance metric from the equivalence constraints [39]. Its training is a one-pass process and doesn't need any iteration.

Besides the pairwise constrained metric learning methods, some other methods learn the metric based on the triplet constraint. It makes each sample to be close to its similar sample and far from its dissimilar sample. Large Margin Nearest Neighbor (LMNN) learns the distance metric by a convex problem with the triplet constraints [40]. It separates the similar and dissimilar neighbors of each sample by a large margin. Shen et al. propose the BoostMetric [41], MetricBoost [42] and FrobMetric [43] methods. They parameterize the distance metric as the linear combination of rank-one matrices, and learn the combination parameters based on triplet constraints.

The construction of pairwise or triplet constraints is crucial to the performance of metric learning methods. Among the existing metric learning methods, many methods construct the constraints by randomly selecting the pairs or triplets. Wang et al. [7] propose a nearest neighbor strategy to construct the pairs and triplets, and proved that this strategy can lead to higher recognition accuracy than the random selection strategy. However, the pairs and triplets are fixed in their training process. In our work, we train the model for many cycles, and we dynamically update the pairwise constraints by the nearest neighbor strategy in each cycle as shown in Figure 1; where **T** is the distance threshold. So our proposed method can incorporate more pairwise constraints in training.

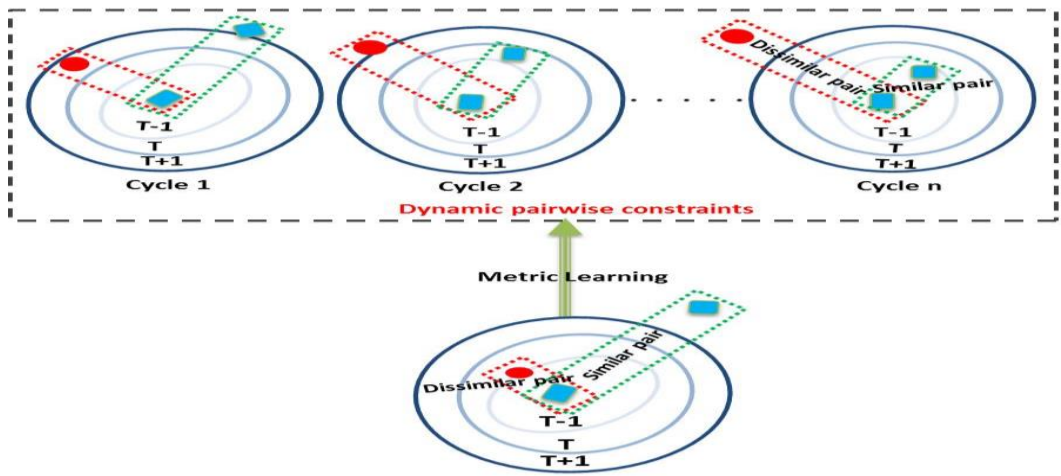

**Figure 1**. Diagram of dynamic pairwise constraints

## 3 Proposed Method

In this section, we propose the ear recognition based on LogDet divergence (ERLD) method for ear recognition. It first resize all images to fixed size of 100 * 100 pixels and apply histogram equalization to the resized image, then extracts the LPQ, HOG, and Gabor features [44] from the ear images, and then adopts the discriminant correlation analysis (DCA) algorithm [45] to reduce the feature dimension. Finally it learns a Mahalanobis distance metric based on the extracted feature. The sketch of the proposed method is illustrated in Figure 2.

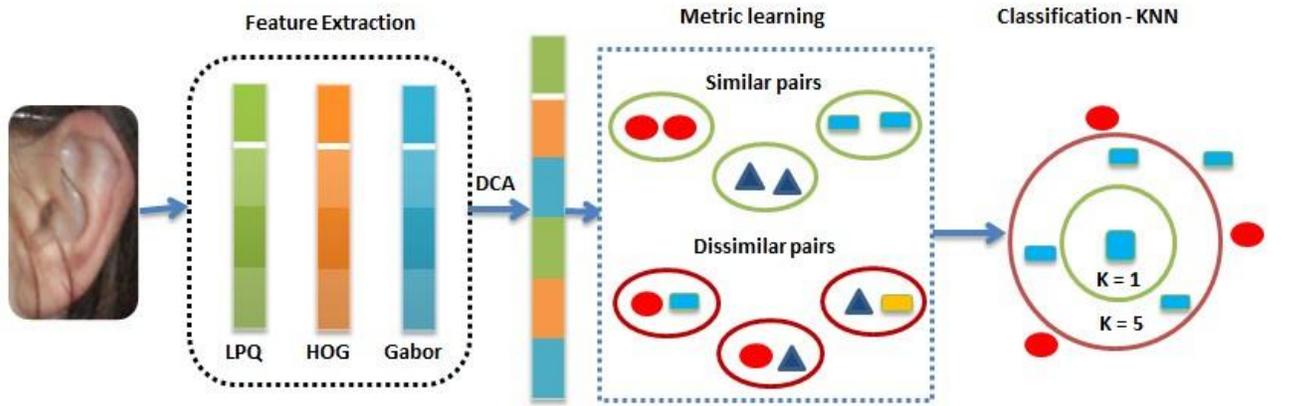

**Figure 2**. The sketch of the proposed ear recognition method

### 3.1 Problem formulation

Given $n$ training samples $\{(\mathbf{x}_i, y_i) | i = 1, 2, \cdots, n\}$, where $\mathbf{x}_i \in \mathrm{R}^d$, the squared Mahalanobis distance between $\mathbf{x}_i$ and $\mathbf{x}_j$ is defined as

$$D_\mathbf{A}(\mathbf{x}_i, \mathbf{x}_j) = (\mathbf{x}_i - \mathbf{x}_j)^T \mathbf{A} (\mathbf{x}_i - \mathbf{x}_j) \qquad (1)$$

where **A** is the distance metric which is a positive semidefinite matrix.

Denote by $S = \{(\mathbf{x}_i, \mathbf{x}_j) | \mathbf{x}_i \text{ and } \mathbf{x}_j \text{ are from the same person}\}$ the set of similar pairs and $D = \{(\mathbf{x}_i, \mathbf{x}_j) | \mathbf{x}_i \text{ and } \mathbf{x}_j \text{ are from different persons}\}$ the set of dissimilar pairs. We hope the pairs satisfy the following constraints

$$D_\mathbf{A}(\mathbf{x}_i, \mathbf{x}_j) \begin{cases} \leq U, & \forall (\mathbf{x}_i, \mathbf{x}_j) \in S \\ \geq L, & \forall (\mathbf{x}_i, \mathbf{x}_j) \in D \end{cases} \quad (2)$$

where $U$ and $L$ are the upper and lower thresholds.

Similar with ITML, we formulate the objective function as that of minimizing the LogDet divergence function to make the learned distance metric $\mathbf{A}$ to be close to the given prior distance metric $\mathbf{P}$. Thus the problem is formulated as follows

$$\begin{aligned} \min_{A,\xi} \quad & \mathbf{B}_{ld}(\mathbf{A},\mathbf{P}) + \gamma \mathbf{B}_{ld}(\xi_0, \xi) \\ s.t \quad & D_A(x_i, x_i) \leq \xi_{ij}, \quad \forall (x_i, x_j) \in S \\ & D_A(x_i, x_i) \geq \xi_{ij}, \quad \forall (x_i, x_j) \in D \\ & A \geq 0, \; \xi_{ij} \geq 0 \end{aligned} \quad (3)$$

where $\mathbf{B}_{ld}(\mathbf{A},\mathbf{P}) = tr(\mathbf{AP}^{-1}) - \log\det(\mathbf{AP}^{-1}) - d$ is the LogDet divergence function between two matrices $\mathbf{A}$ and $\mathbf{P}$ [46], [6], $tr(\bullet)$ is the trace of the matrix, and $d$ is the dimension of training samples.

## 3.2 Training Algorithm

### 3.2.1 Construction of the pairwise constraints

In the proposed method, we initialize the similar and dissimilar pair sets in the first cycle, and then update the pair sets in each of the rest cycles. Denote by $S_k$, $D_k$ and $\mathbf{A}_k$ the sets of similar and dissimilar pairs, and the learned distance metric in the $k$th cycle, respectively. In the first cycle, we initialize $\mathbf{P}$ as the identity matrix, compute the Euclidean distance of every two training samples, and then find the nearest similar and dissimilar neighbors of each sample to initialize $S_0$ and $D_0$, respectively. In the $k$th cycle, we use $\mathbf{A}_{k-1}$ as the prior distance metric to compute the Mahalanobis distances of every two training samples. Based on the distances, we find the nearest similar and dissimilar neighbors of each sample to set $S_k$ and $D_k$, respectively.

### 3.2.2 Optimization of the distance metric

In each training cycle, we set the prior distance metric $\mathbf{P}$ as $\mathbf{A}_{k-1}$, and solve the problem (4) with the training pairs $S_k$ and $D_k$. Following [6], we initialize $\mathbf{A}_k$ as $\mathbf{A}_{k-1}$, and learn the distance metric by repeatedly compute the Bregman projections as follows:

$$\mathbf{A}_k = \mathbf{A}_k + \beta_{ij} \mathbf{A}_k \left(\mathbf{x}_i - \mathbf{x}_j\right)\left(\mathbf{x}_i - \mathbf{x}_j\right)^T \mathbf{A}_k$$

where $(\mathbf{x}_i, \mathbf{x}_j)$ is a training pair in $S_k$ or $D_k$, and $\beta_{ij}$ is the Lagrange multiplier corresponding to pair $(\mathbf{x}_i, \mathbf{x}_j)$.

### 3.2.3 The algorithm of the proposed method

As described in Sections 3.2.1 and 3.2.2, we summarize the algorithm of the proposed method as **Algorithm 1**.

---

**Algorithm 1** The training algorithm of the proposed method

---

**Input:** Training set $\{(\mathbf{x}_i, y_i) | i = 1, 2, \cdots, n\}$, cycle number $m$,

**Output:** Learned distance metric **A**.
1. Initialize the prior distance metric $\mathbf{A}_0$ as the identity matrix.
2. **For** $k = 1$ to $m$
2.1. $S_k \leftarrow \varnothing$, $D_k \leftarrow \varnothing$
2.2. Compute the distances of every two training samples with the distance metric $\mathbf{A}_{k-1}$.
2.3. **For each** training sample $\mathbf{x}_i$
2.3.1. Find the nearest similar and dissimilar neighbors of $\mathbf{x}_i$ as $\mathbf{x}_p$ and $\mathbf{x}_q$.
2.3.2. $S_k \leftarrow S_k \cup \{(\mathbf{x}_i, \mathbf{x}_p)\}$, $D_k \leftarrow D_k \cup \{(\mathbf{x}_i, \mathbf{x}_q)\}$.
2.4. End for
2.5. $\mathbf{A}_k \leftarrow \mathbf{A}_{k-1}$
2.5. **Repeat**
2.5.1. Pick a pair $(\mathbf{x}_i, \mathbf{x}_j)$ in $S_k \cup D_k$
2.5.2. $p \leftarrow (\mathbf{x}_i - \mathbf{x}_j)^T \mathbf{A} (\mathbf{x}_i - \mathbf{x}_j)$.
2.5.3. $\delta \leftarrow 1$ if $(\mathbf{x}_i, \mathbf{x}_j) \in S_k$, and $\delta \leftarrow -1$ if $(\mathbf{x}_i, \mathbf{x}_j) \in D_k$.
2.5.4. $\alpha \leftarrow \min\left(\lambda_{ij}, \delta/2\left(1/p - \gamma/\xi_{ij}\right)\right)$
2.5.5. $\beta \leftarrow \delta\alpha/(1 - \delta\alpha p)$
2.5.6. $\xi_{ij} \leftarrow \gamma\xi_{ij}/(\gamma + \delta\alpha\xi_{ij})$
2.5.7. $\lambda_{ij} \leftarrow \lambda_{ij} - \alpha$
2.5.8. $\mathbf{A}_k \leftarrow \mathbf{A}_k + \beta \mathbf{A}_k (\mathbf{x}_i - \mathbf{x}_j)(\mathbf{x}_i - \mathbf{x}_j)^T \mathbf{A}_k$
2.6. **Until** convergence
2.7. $k \leftarrow k + 1$
3. End for
4. **Return** $\mathbf{A}_m$

---

## 4 Datasets and Experimental Results

To evaluate our proposed approach, the experiments are conducted on three ear databases, i.e., West Pommeranian University of Technology (WPUT) [47], the University of Science and Technology Beijing II (USTB II) [8] and Mathematical Analysis of Images (AMI) [48] databases.

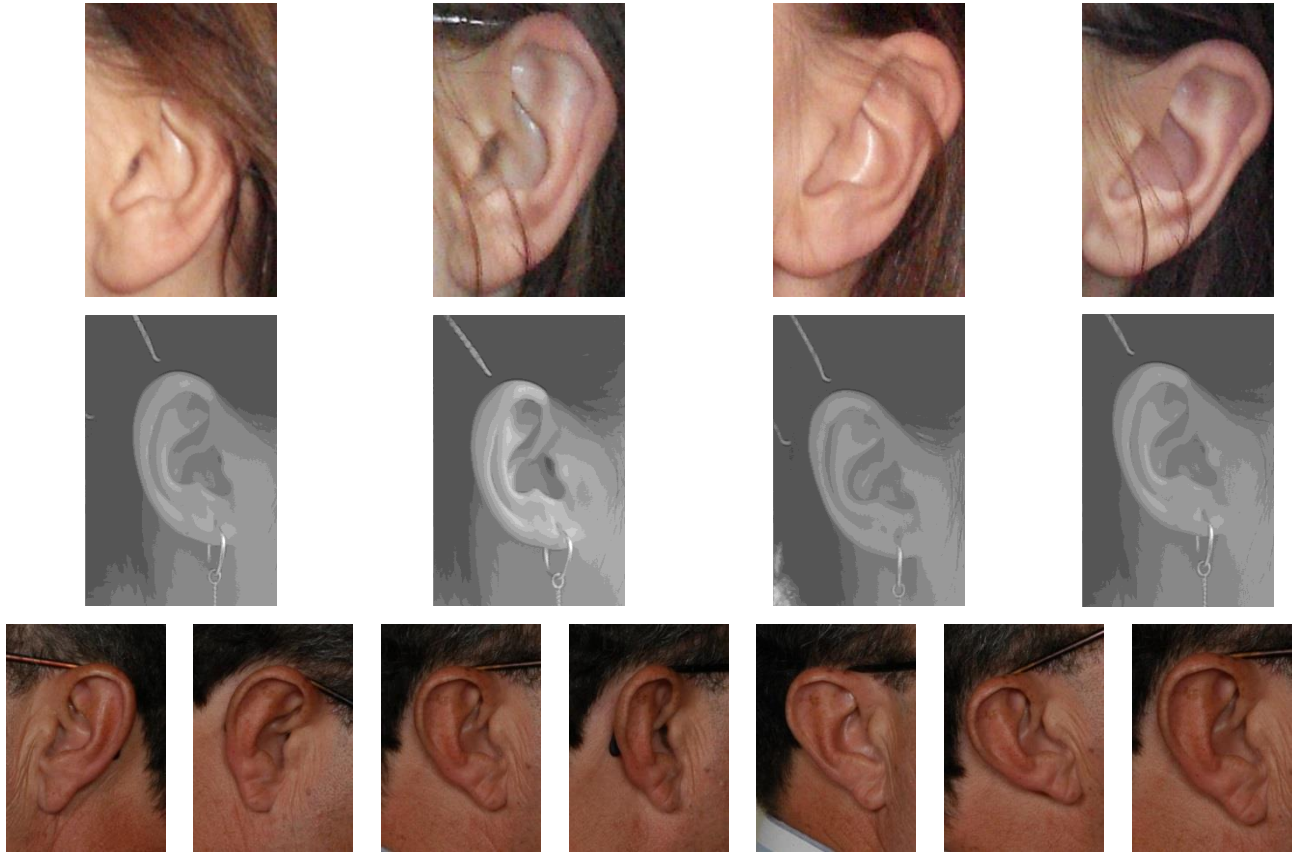

**Figure 3.** Original ear images for one subject from WPUT (up row), USTB II (middle row) and AMI (bottom row).

The WPUT [47] ear database was introduced in 2010 and consists of 3345 images of 475 persons with 1388 duplicates, among which each person has 4~10 images; we are used only 4 images for each subject. The images are taken from men and women, and under different indoor lightning conditions and head rotation angles ranging from approximately 90° for profile to 75°, and occlusion include earrings, hat, tattoos, etc. The USTB II database [8] contains 308 images of 77 persons, which are taken under different illumination and camera views. Each person has 4 images. The first image is the frontal ear image under standard illumination, the second and the third images are taken with +30° and −30° rotations respectively, and the fourth image is taken under weak illumination. Fig. 3 shows the original ear images for one subject from these two databases. The AMI ear database has 700 images from 100 persons, all of subjects in the age range of 19~65 years. AMI ear images are collected from students, teachers and staff at Universidad de Las Palmas de Gran Canaria (ULPGC), Las Palmas, Spain, and taken in an indoor environment. Each person has 7 images; five of them were right side profile (right ear) and the sixth image of right profile was taken but with a different camera, last image was taken from a left side profile (left ear).

**Table 1.** The recognition rate of different methods in WPUT database

|  | k=1 | k=2 | k=3 | k=4 | k=5 |
|---|---|---|---|---|---|
| Euclidean distance | 95.37 | 95.37 | 95.37 | 95.37 | 95.37 |
| ITML [6] | 95.63 | 95.85 | 95.37 | 94.90 | 94.90 |
| LMNN [40] | 94.21 | 94.21 | 85.26 | 85.26 | 70.26 |
| LDML[36] | 94.85 | 94.90 | 89.13 | 84.76 | 80.94 |
| LDMLT [49] | 94.58 | 94.63 | 94.63 | 94.37 | 94.16 |
| Ours | **98.74** | **98.74** | **98.84** | **98.79** | **98.58** |

**Table 2.** The recognition rate of different methods in USTB II database

|  | k=1 | k=2 | k=3 | k=4 | k=5 |
|---|---|---|---|---|---|
| Euclidean distance | 95.78 | 95.78 | 95.78 | 95.78 | 95.78 |
| ITML [6] | 96.10 | 88.96 | 85.06 | 82.14 | 79.22 |
| LMNN [40] | 96.10 | 96.10 | 87.33 | 87.33 | 69.15 |
| LDML [36] | 97.14 | 97.14 | 96.33 | 95.52 | 95.00 |
| LDMLT [49] | 95.65 | 95.65 | 95.45 | 95.36 | 95.32 |
| Ours | **98.70** | **98.70** | **98.70** | **98.38** | **98.38** |

We compare our proposed method with the state-of-the-art metric learning algorithms, i.e., ITML [6], LMNN [40], LMDT [49] and LDMLT [49]. We report the recognition accuracy using the *k*-nearest neighbor (*k*-NN) classifier with $k = 1, 2, ... 5$ via three-fold cross validation in Table 1, Table 2 and Table 3.

**Table 3.** The recognition rate of different methods in AMI database

|  | k=1 | k=2 | k=3 | k=4 | k=5 |
|---|---|---|---|---|---|
| Euclidean distance | 95.14 | 95.14 | 95.86 | 96.00 | 96.43 |
| ITML [6] | 96.71 | 93.71 | 97 | 96.57 | 89.86 |
| LMNN [40] | **98.89** | **98.89** | **98.29** | **98.29** | 96.67 |
| LDML [36] | 97.96 | 97.96 | 98.19 | 97.71 | 97.67 |
| LDMLT [49] | 97.07 | 97.07 | 97.43 | 97.70 | 97.59 |
| Ours | 97.31 | 97.31 | 97.76 | 97.97 | **98.16** |

We can see that our proposed method can achieve better performances than the other competing methods in the WPUT and USTB II database, even if the WPUT and USTB II database is small-scale. In the AMI database, the recognition rates of our proposed method are slightly lower than those of LMNN and LDML. That's because LDML and LMNN are based on triplet constraints, while our proposed method is based on pairwise constraints. However, we also compare the training time of our proposed method and the other state-of-the-art metric learning methods in Table 4. We can see that the training time of our proposed method is much shorter than the other metric learning methods. We analyzed the computational complexity of our proposed method in training. From Eq. (3), the computational complexity in each iteration is $O(d^2)$. As there are $m$ cycles and $O(n)$ pairs in each cycle, the computational complexity of our proposed method is $O(mnd^2)$.

**Table 4.** Comparison of the training time of different metric learning methods

| Methods / Databases | ITML [6] | LMNN [40] | LDML [36] | LDMLT [49] | Our method (ERLD) |
|---|---|---|---|---|---|
| AMI | 54.91 sec | 82.58 sec | 1.86 sec | 3.48 sec | **0.54 sec** |

# 5 Conclusion

This paper proposes a novel pairwise constrained metric learning method for ear recognition. In this method, we update the pairwise constraint using the nearest neighbor strategy in each training cycle, and learn the distance metric via minimizing the LogDet divergence of the learned metric and prior metric. This method can incorporate more pairwise constraints to learn the distance metric, which can improve the recognition performance with efficient training time. The experimental results on the AMI, WPUT and USTB II databases show that our proposed method can achieve favorable recognition accuracy, and its training time is much faster than the other competing methods. In the future, we will investigate to apply this strategy to dynamically build the triplet constraint and propose more metric learning methods for ear recognition.

## Acknowledgments

This work was supported in part by the co-operation between Higher Education Commission of Egypt and Chinese Government and the National Natural Science Foundation of China under Grant No. 61271093 and 61471146.